\documentclass[letterpaper, 10 pt, conference]{ieeeconf}  

\IEEEoverridecommandlockouts                              

\usepackage{amsthm}
\usepackage{graphicx}
\usepackage{caption}
\usepackage{subcaption}
\usepackage[linesnumbered,ruled,vlined]{algorithm2e}
\usepackage{algorithmicx}
\usepackage{amsmath}
\theoremstyle{plain}
\usepackage{mathtools}
\usepackage{epsfig}
\usepackage{amssymb}
\usepackage{epstopdf}
\usepackage{multicol}
\usepackage{multirow}
\usepackage{array}
\usepackage{url}
\usepackage{color}
\usepackage{float}
\usepackage{wrapfig}
\usepackage[para]{footmisc}

\title{\LARGE \bf
Motion Planning Networks  
}

\author{Ahmed H. Qureshi, Anthony Simeonov, Mayur J. Bency and Michael C. Yip
\thanks{A. H. Qureshi$^1$, A. Simeonov$^2$, M. J. Bency$^1$, and M. C. Yip$^{1,2}$ are with $(1)$ Department of Electrical and Computer Engineering; $(2)$ Department of Mechanical and Aerospace Engineering; University of California San Diego, La Jolla, CA 92093 USA. {\tt\small \{a1qureshi, asimeono, mbency, yip\}@ucsd.edu}}%
}

\begin{document}

\maketitle
\thispagestyle{empty}
\pagestyle{empty}

\begin{abstract}
Fast and efficient motion planning algorithms are crucial for many state-of-the-art robotics applications such as self-driving cars. Existing motion planning methods become ineffective as their computational complexity increases exponentially with the dimensionality of the motion planning problem. To address this issue, we present Motion Planning Networks (MPNet), a neural network-based novel planning algorithm. The proposed method encodes the given workspaces directly from a point cloud measurement and generates the end-to-end collision-free paths for the given start and goal configurations. We evaluate MPNet on various 2D and 3D environments including the planning of a 7 DOF Baxter robot manipulator. The results show that MPNet is not only consistently computationally efficient in all environments but also generalizes to completely unseen environments. The results also show that the computation time of MPNet consistently remains less than 1 second in all presented experiments, which is significantly lower than existing state-of-the-art motion planning algorithms.
\end{abstract}

\IEEEpeerreviewmaketitle

\section{Introduction}
Robotic motion planning aims to compute a collision-free path for the given start and goal configurations \cite{lavalle2006planning}. As motion planning algorithms are necessary for solving a variety of complicated, high-dimensional problems ranging from autonomous driving  \cite{lozano2012autonomous} to space exploration \cite{volpe2003rover}, there arises a critical, unmet need for computationally tractable, real-time algorithms. The quest for developing computationally efficient motion planning methods has led to the development of various sampling-based motion planning (SMP) algorithms such as Rapidly-exploring Random Trees (RRT) \cite{lavalle1998rapidly}, optimal Rapidly-exploring Random Trees (RRT*) \cite{karaman2011sampling}, Potentially guided-RRT* (P-RRT*) \cite{qureshi2016potential} and their bi-directional variants \cite{qureshi2015intelligent,tahir2018potentially}. Despite previous efforts to design fast, efficient planning algorithms, the current state-of-the-art struggles to offer methods which scale to the high-dimensional setting that is common in many real-world applications. 

To address the above-mentioned challenges, we propose a Deep Neural Network (DNN) based iterative motion planning algorithm, called MPNet (Motion Planning Networks) that efficiently scales to high-dimensional problems. MPNet consists of two components: an encoder network and a planning network. The encoder network learns to encode a point cloud of the obstacles into a latent space. The planning network learns to predict the robot configuration at time step $t+1$ given the robot configuration at time $t$, goal configuration, and the latent-space encoding of the obstacle space. Once trained, MPNet can be used in conjunction with our novel bi-directional iterative algorithm to generate feasible trajectories.
\begin{figure}
\vspace*{0.05in}
    \centering
       \includegraphics[width=8.5cm]{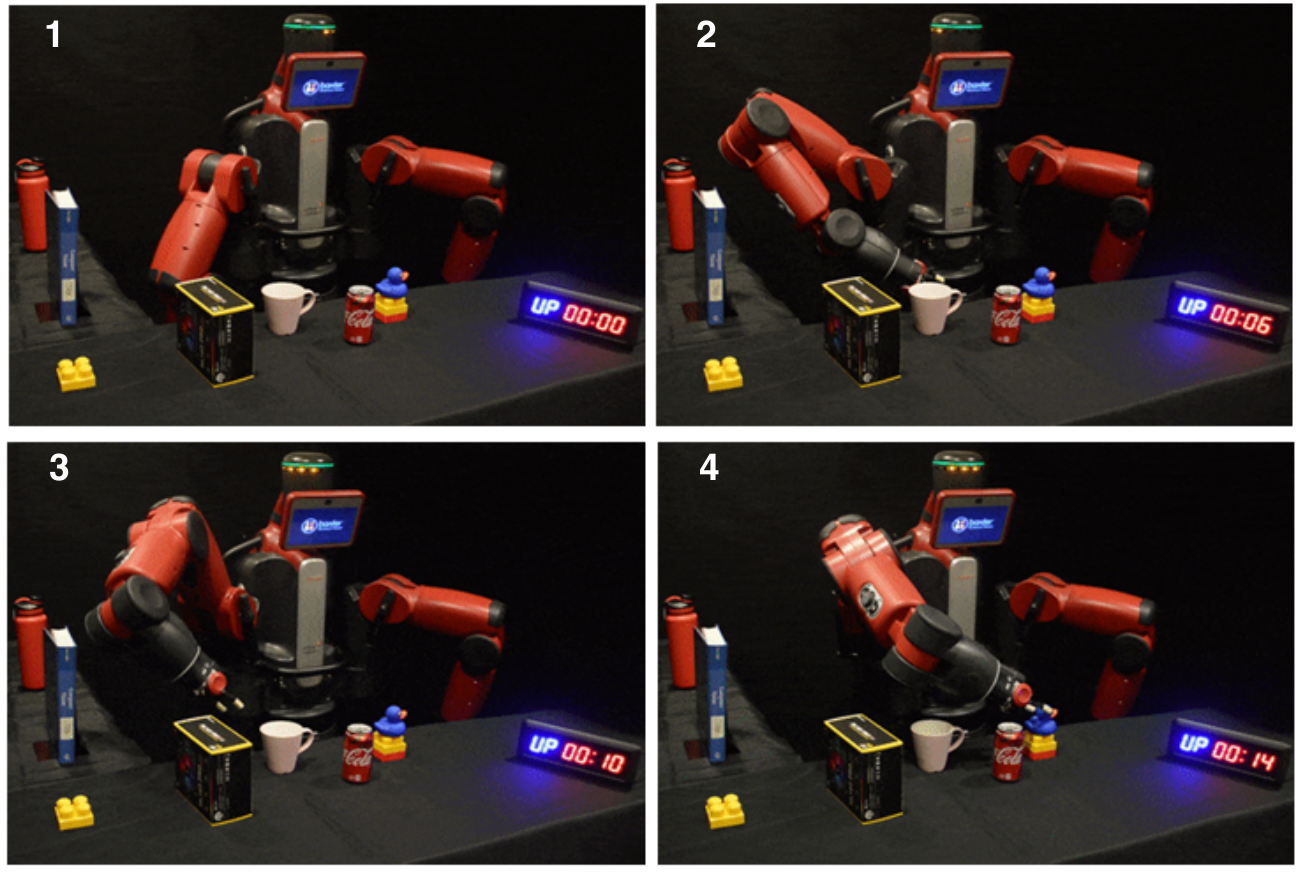}
    \caption{MPNet planned motion for a 7 DOF Baxter robot manipulator. The path profile followed by the robot from initial to target configuration is shown through frames 1-4. The stopwatch in the images show the execution time. In this particular case, MPNet took less than 1 second whereas BIT* \cite{gammell2015batch} took 3.1 minutes on average to find a feasible path solution of comparable euclidean cost.}\label{baxter}
\vspace*{-0.15in}\end{figure}  
We evaluate MPNet on a large test dataset including multiple planning problems such as the planning of a point-mass robot, rigid-body, and 7 DOF Baxter robot manipulator in various 2D and 3D environments.
As neural networks do not provide theoretical guarantees on their performance, we also propose a hybrid algorithm which combines MPNet with any existing classical planning algorithm, in our case RRT*. The hybrid planning technique demonstrates a 100\% success rate consistently over all tested environments while retaining the computational gains. Our results indicate that MPNet generalizes very well, not only to unseen start and goal configurations within workspaces which were used in training, but also to new workspaces which the algorithm has never seen. 
\section{Related Work}
Research into developing neural network-based motion planners first gained traction in the early 1990s but faded away due to computational complexity of training deep neural networks \cite{schmidhuber2015deep}. However, recent developments in Deep Learning (DL) have allowed researchers to apply various DL architectures to robotic control and planning. 

An active area of research within robotic control and planning is Deep Reinforcement Learning (DRL). For instance, \cite{levine2016end} shows how to train a robot to learn visuomotor policies to perform various tasks such as screwing a bottle cap, or inserting a peg. Although DRL is a promising framework, it extensively relies on exploration through interaction with the environment, thus making it difficult to train for many real-world robotic applications. A recent work, Value Iteration Networks (VIN) \cite{tamar2016value} emulates value iteration by leveraging recurrent convolutional neural networks and max-pooling. However, in addition to limitations inherited from underlying DRL framework, VIN has only been evaluated on simple toy problems, and it is not clear how VIN could extend beyond such environments.

Imitation learning is another emerging field in which the models are trained from expert demonstrations. Many interesting problems have been addressed through imitation learning \cite{bojarski2016end, calinon2010learning,rahmatizadeh2016learning}. A recent method \cite{ichter2018learning} uses deep neural networks trained via imitation to adaptively sample the configuration space for SMP methods. Our proposed method also learns through imitation but unlike \cite{ichter2018learning}, it provides a complete feasible motion plan for a robot to follow.

Another recent and relevant method is the Lightning Framework \cite{berenson2012robot}, which is composed of two modules. The first module performs path planning using any traditional motion planner. The second module maintains a lookup table which caches old paths generated by the first module. For new planning problems, the Lightning Framework retrieves the closest path from a lookup table and repairs it using a traditional motion planner. This approach demonstrates superior performance compared to conventional planning methods. However, not only are lookup tables memory inefficient, they also are incapable of generalizing to new environments.
  
\section{Problem Definition}

This section describes the notations used in this paper and formally defines the motion planning problem addressed by the proposed method.  

Let $Q$ be an ordered list of length $N \in \mathbb{N}$, then a sequence $\{q_i=Q(i)\}_{i \in N}$ is a mapping from $i \in \mathbb{N}$ to the $i$-th element of $Q$. Moreover, for the algorithms described in this paper, $Q(\mathrm{end})$ and $Q.\mathrm{length}()$ give the last element and the number of elements in a set $Q$, respectively. Let $X \subset \mathbb{R}^d$ be a given state space, where $d \in \mathbb{N}$ is the dimensionality of the state space. The workspace dimensionality is indicated by $d_w \in \mathbb{N}$. The obstacle and obstacle-free state spaces are defined as $X_\mathrm{obs} \subset X$ and $X_\mathrm{free}= X \backslash X_\mathrm{obs}$, respectively. Let the initial state be $\boldsymbol{x_\mathrm{init}} \in X_\mathrm{free}$, and goal region be $X_\mathrm{goal} \subset X_\mathrm{free}$. Let an ordered list $\tau$ be a path having positive scalar length. A solution path $\tau$ to the motion planning problem is feasible if it connects $\boldsymbol{x_\mathrm{init}}$ and $ \boldsymbol{x} \in X_\mathrm{goal}$, i.e. $\tau(0)=\boldsymbol{x_\mathrm{init}}$ and $\tau(\mathrm{end}) \in X_\mathrm{goal}$, and lies entirely in the obstacle-free space $X_\mathrm{free}$. The proposed work addresses the feasibility problem of motion planning.
\section{MPNet: A Neural Motion Planner}
This section introduces our proposed model, MPNet\footnote{Supplementary material including implementation parameters and project videos are available at https://sites.google.com/view/mpnet/home.} (see Fig. \ref{MPNet}). MPNet is a neural network based motion planner comprised of two phases: (A) offline training of the neural models, and (B) online path generation.
\subsection{Offline Training}
Our proposed method uses two neural models to solve the motion planning problem. The first model is an encoder network which embeds the obstacles point cloud, corresponding to a point cloud representing $X_{\mathrm{obs}}$, into a latent space (see Fig. \ref{MPNet}(a)). The second model is a planning network (Pnet) which learns to do motion planning for the given obstacle embedding, and start and goal configurations of the robot (see Fig. \ref{MPNet}(b)).
\subsubsection{Encoder Network}
The encoder network (Enet) embeds the obstacles point cloud into a feature space $\boldsymbol{Z} \in \mathbb{R}^m$ with dimensionality $m \in \mathbb{N}$. Enet can be trained either using encoder-decoder architecture with a reconstruction loss or in an end-to-end fashion with the Pnet (described below). For encoder-decoder training, we found that the contrative autoencoders (CAE) \cite{rifai2011contractive} learns robust and invariant feature space required for planning and genalization to unseen workspaces. The reconstruction loss of CAE is defined as: 
\begin{equation}\vspace*{-0.1in}
L_\mathrm{AE}\big(\boldsymbol{\theta}^e,\boldsymbol{\theta}^d\big)= \cfrac{1}{N_\mathrm{obs}}\sum_{\boldsymbol{x}\in D_\mathrm{obs}}||\boldsymbol{x}-\hat{\boldsymbol{x}}||^2  + \lambda \sum_{ij} (\theta^e_{ij})^2
\end{equation}
where $\boldsymbol{\theta}^e$ are the parameters of encoder, $\boldsymbol{\theta}^d$ are the parameters of decoder, $\lambda$ is a penalizing coefficient, $D_\mathrm{obs}$ is a dataset of point clouds $\boldsymbol{x} \in X_\mathrm{obs}$ from $N_\mathrm{obs} \in \mathbb{N}$ different workspaces, and $\hat{\boldsymbol{x}}$ is the point cloud reconstructed by the decoder. 
%
\subsubsection{Planning Network}
We use a feed-forward deep neural network, parameterized by $\boldsymbol{\theta}$, to perform planning. Given the obstacles encoding $\boldsymbol{Z}$, current state $\boldsymbol{x}_t$ and the goal state $\boldsymbol{x}_T$, Pnet predicts the next state $\hat{\boldsymbol{x}}_{t+1} \in X_\mathrm{free}$ which would lead a robot closer to the goal region, i.e., $\hat{\boldsymbol{x}}_{t+1}=\mathrm{Pnet}((\boldsymbol{x}_t,\boldsymbol{x}_T,\boldsymbol{Z});\boldsymbol{\theta})$

To train Pnet, any planner or human expert can provide feasible, near-optimal paths as expert demonstrations. We assume the paths given by the expert demonstrator are in a form of a tuple, $\tau^*=\{\boldsymbol{x}_0,\boldsymbol{x}_1,\cdots,\boldsymbol{x}_T\}$, of feasible states that connect the start and goal configurations so that the connected path lies entirely in $X_\mathrm{free}$. The training objective for the Pnet is to minimize the mean-squared-error (MSE) loss between the predicted states $\hat{\boldsymbol{x}}_{t+1}$ and the actual states $\boldsymbol{x}_{t+1}$ given by the expert data, formalized as:
\begin{equation}\vspace*{-0.1in}
L_\mathrm{Pnet}(\boldsymbol{\theta})=\cfrac{1}{N_p} \sum^{\hat{N}}_j \sum^{T-1}_{i=0} ||\hat{\boldsymbol{x}}_{j,i+1}-\boldsymbol{x}_{j,i+1}||^2,
\end{equation}
where $N_p \in \mathbb{N}$ is the averaging term corresponding to the total number of paths, $\hat{N} \in \mathbb{N}$, in the training dataset times the path lengths. 
\begin{figure*}
    \centering
   \begin{subfigure}[b]{0.31\textwidth}
        \includegraphics[width=5.75cm]{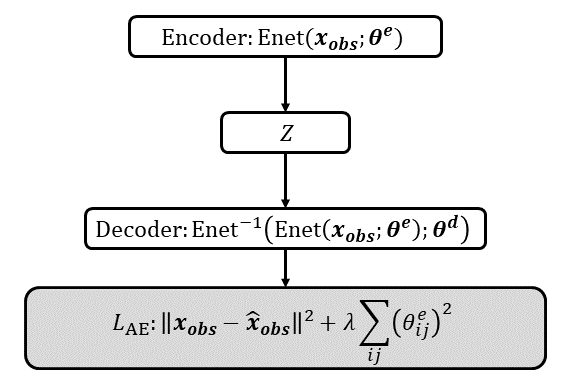}
       \caption{Offline: Encoder Network}
    \end{subfigure}
     \begin{subfigure}[b]{0.31\textwidth}
        \includegraphics[width=5.70cm]{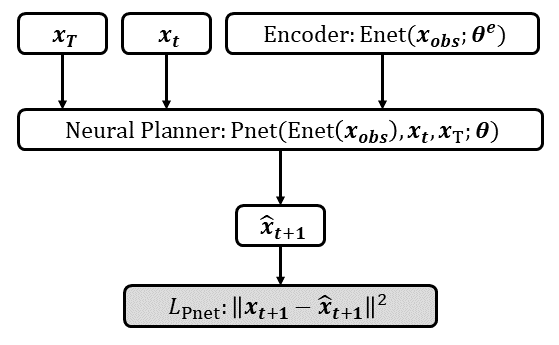}
        \caption{Offline: Planning Network}
    \end{subfigure}
    \begin{subfigure}[b]{0.31\textwidth}
        \includegraphics[width=5.92cm]{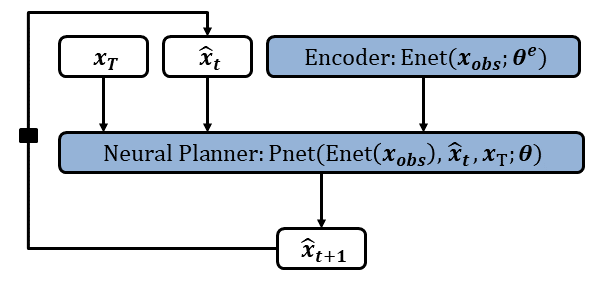}
        \caption{Online: Neural Planner}
    \end{subfigure}
    \caption{The offline and online phases of MPNet. The grey shaded blocks indicate the training objectives. The blue blocks represents frozen modules that do not undergo any training. }\label{MPNet}
\vspace*{-0.15in}\end{figure*}

\subsection{Online Path Planning}
The online phase exploits the neural models from the offline phase to do motion planning in cluttered and complex environments. The overall flow of information between Enet and Pnet is shown in Fig. \ref{MPNet}(c). To generate end-to-end feasible paths connecting the start and goal states, we propose a novel incremental bidirectional path generation heuristic. Algorithm 1 presents the overall path generation procedure and its constituent functions are described below.
\begin{algorithm}[b]
\DontPrintSemicolon 
$\boldsymbol{Z} \gets \mathrm{Enet}(\boldsymbol{x_\mathrm{obs}})$\;
$\tau \gets \mathrm{NeuralPlanner}(\boldsymbol{x_\mathrm{init}},\boldsymbol{x_\mathrm{goal}}, \boldsymbol{Z});$\;
\If{$\tau$}
   {

$\tau \gets \mathrm{LazyStatesContraction}(\tau)$\;
 \If{$\mathrm{IsFeasible}(\tau)$}
   {
      \Return$\tau$\
   }
   \Else
   {
   	$\tau_\mathrm{new} \gets \mathrm{Replanning(\tau,\boldsymbol{Z})}$\;
   	$\tau_\mathrm{new} \gets \mathrm{LazyStatesContraction}(\tau_\mathrm{new})$\;
 \If{$\mathrm{IsFeasible}(\tau_\mathrm{new})$}
   {
      \Return$\tau_\mathrm{new}$\;
   }

   \Return$\varnothing$\;
   }
   
   }

\caption{MPNet($\boldsymbol{x_\mathrm{init}},\boldsymbol{x_\mathrm{goal}}, \boldsymbol{x_\mathrm{obs}}$)}
\label{algo:NPG}
\end{algorithm} 
\subsubsection{Enet}
The encoder network $\mathrm{Enet}(\boldsymbol{x_\mathrm{obs}})$, trained during the offline phase, is used to encode the obstacles point cloud $\boldsymbol{x_\mathrm{obs}} \in X_\mathrm{obs}$ into a latent space $\boldsymbol{Z} \in \mathbb{R}^m$. 
\subsubsection{Pnet}
Pnet is a feed-forward neural network from the offline phase which takes $\boldsymbol{Z}$, current state $\boldsymbol{x}_t$, goal state $\boldsymbol{x}_T$ and predicts the next state of the robot $\hat{\boldsymbol{x}}_{t+1}$. To inculcate stochasticity into the Pnet, some of the hidden units in each of its hidden layer were dropped out with a probability $p:[0,1] \in \mathbb{R}$. The merit of adding the stochasticity during the online path generation are presented in the discussion section. 
\subsubsection{Lazy States Contraction (LSC)}
Given a path $\tau=\{\boldsymbol{x}_0,\boldsymbol{x}_1, \cdots, \boldsymbol{x}_T\}$, the LSC algorithm connects the directly connectable non-consecutive states, i.e., $\boldsymbol{x}_i$ and $\boldsymbol{x}_{>i+1}$, and removes the intermediate/lazy states. 
\subsubsection{Steering}
The $\mathrm{steerTo}$ function takes two states as an input and checks either a straight trajectory connecting the given two states lies entirely in collision-free space $X_\mathrm{free}$ or not. The steering is done from $\boldsymbol{x}_1$ to $\boldsymbol{x}_2$ in small, discrete steps and can be summarized as $\tau(\delta) = (1 - \delta) \boldsymbol{x}_1 + \delta \boldsymbol{x}_2; \forall\delta \in [0, 1]$. The discrete step size could be a fixed number or can be adapted for different parts of the algorithm.  

\subsubsection{isFeasible}
Given a path $\tau=\{\boldsymbol{x}_0,\boldsymbol{x}_1, \cdots, \boldsymbol{x}_T\}$, this procedure checks either the end-to-end path, formed by connecting the consecutive states in $\tau$, lies entirely in $X_\mathrm{free}$ or not. 
\begin{algorithm}[b]
\DontPrintSemicolon 
$\tau^\mathrm{a} \gets \{\boldsymbol{x_\mathrm{start}}\}; \tau^\mathrm{b} \gets \{\boldsymbol{x_\mathrm{goal}}\};$\;

$\tau\gets \varnothing;$\;
$\mathrm{Reached} \gets \mathrm{False};$\;
\For{$i \gets 0$ \textbf{to} $N$} {
	
	$\boldsymbol{x_\mathrm{new}} \gets \mathrm{Pnet}\big(\boldsymbol{Z},\tau^\mathrm{a}(\mathrm{end}),\tau^\mathrm{b}(\mathrm{end})\big)$\;
	$\tau^\mathrm{a} \gets \tau^\mathrm{a} \cup \{\boldsymbol{x_\mathrm{new}}\}$\;
	$\mathrm{Connect} \gets \mathrm{steerTo}\big(\tau^\mathrm{a}(\mathrm{end}),\tau^\mathrm{b}(\mathrm{end})\big)$\;
   \If{$\mathrm{Connect}$}
   {
   	  $\tau \gets \mathrm{concatenate}(\tau^\mathrm{a},\tau^\mathrm{b})$\;
      \Return{$\tau$}\;
   }
   $\mathrm{SWAP}(\tau^\mathrm{a},\tau^\mathrm{b})$\;
   }

\Return{$\varnothing$}\;
\caption{NeuralPlanner($\boldsymbol{x_\mathrm{start}}, \boldsymbol{x_\mathrm{goal}}, \boldsymbol{Z}$)}
\label{algo:NP}
\end{algorithm} 
\subsubsection{Neural Planner}
This is an incremental bidirectional path generation heuristic (see Algorithm 2 for the outline). It takes the obstacles' representation, $Z$, as well as the start and goal states as an input, and outputs a path connecting the two given states. The sets $\tau^a$ and $\tau^b$ correspond to the paths generated from the start and goal states, respectively. The algorithm starts with $\tau^a$, it generates a new state $\boldsymbol{x_\mathrm{new}}$, using Pnet, from start towards the goal (Line 5), and checks if a path from start $\tau^a$ is connectable to the path from a goal $\tau^b$ (Line 7). If paths are connectable, an end-to-end path $\tau$ is returned by concatenating $\tau^a$ and $\tau^b$. However, if paths are not connectable, the roles of $\tau^a$ and $\tau^b$ are swapped (Line 11) and the whole procedure is repeated again. The swap function enables the bidirectional generation of paths, i.e., if at any iteration $i$, path $\tau^a$ is extended then in the next iteration $i+1$, path $\tau^b$ will be extended. This way, two trajectories $\tau^a$ and $\tau^b$ march towards each other which makes this path generation heuristic greedy and fast. 
\begin{algorithm}[t]
\DontPrintSemicolon 
$\tau_\mathrm{new}\gets \varnothing;$\;
\For{$i \gets 0$ \textbf{to} $\tau.\mathrm{length}()$} {
   		\If{$\mathrm{steerTo}(\tau_i,\tau_{i+1})$}
   		{
   			$\tau_\mathrm{new} \gets \tau_\mathrm{new} \cup \{\tau_i,\tau_{i+1}\}  $\;
   		}
   		\Else
   		{
   		$\tau_\mathrm{mini} \gets \mathrm{Replanner}(\tau_i,\tau_{i+1},\boldsymbol{Z});$\;
   		\If{$\tau_\mathrm{mini}$}{
   			$\tau_\mathrm{new} \gets \tau_\mathrm{new} \cup \tau_\mathrm{mini}$\;
   		}
   		\Else
   		{
   		\Return$\varnothing$\; 
   		}
   		}

   	}
   	\Return$\tau_\mathrm{new}$\;
\caption{Replanning($\tau, \boldsymbol{Z}$)}
\label{algo:NPG}
\end{algorithm} 
\subsubsection{Replanning}
This procedure is outlined in the Algorithm 3. It iterates over all the consecutive states $\boldsymbol{x}_i$ and $\boldsymbol{x}_{i+1}$ in a given path $\tau=\{\boldsymbol{x}_0,\boldsymbol{x}_1,\cdots,\boldsymbol{x}_T\}$, and checks if they are connectable or not, where $i=[0,T-1] \subset \mathbb{N}$. If any consecutive states are found not connectable, a new path is generated between those states using one of the following replanning methods (Line 5).  

\textit{ a) Neural Replanning:}
Given a start and goal states together with obstacle space encoding $\boldsymbol{Z}$, this method recursively finds a new path between the two given states. To do so, it starts by finding a coarse path between the given states and then if required, it replans on a finer level by calling itself over the non-connectable consecutive states of the new path. This recursive neural replanning is performed for the fixed number of steps to limit the algorithm within the computational bounds.   

\textit{ b) Hybrid Replanning:} 
This heuristic combines the neural replanning with the classical motion planning methods. It performs the neural replanning for the fixed number of steps. The resulting new path is tested for feasibility. If a path is not feasible, the non-connectable states in the new path are then connected using a classical motion planner.


\section{Implementation details}
This section gives the implementation details of MPNet, for additional details refer to supplementary material. The proposed neural models were implemented in PyTorch \cite{paszke2017automatic}. For environments other than Baxter, the benchmark methods, Informed-RRT* and BIT*, were implemented in Python, and their times were compared against the CPU-time of MPNet. 
The Baxter environments were implemented with MoveIt! \cite{sucan2011} and ROS. In these environments, we use a C++ OMPL \cite{sucan2012the-open-motion-planning-library} implementation of BIT* to compare against a C++ implementation of MPNet. The system used for training and testing has 3.40GHz$\times$ 8 Intel Core i7 processor with 32 GB RAM and GeForce GTX 1080 GPU. The remaining section explains different modules that lead to MPNet.

\subsection{Data Collection}
We generate 110 different workspaces for each presented case, i.e., simple 2D (s2D), rigid-body (rigid), complex 2D (c2D) and 3D (c3D). In each of the workspaces, 5000 collision-free, near-optimal, paths were generated using RRT*. The training dataset comprised of 100 workspaces with 4000 paths in each workspace. For testing, two types of test datasets were created to evaluate the proposed and benchmark methods. The first test dataset, seen-$X_\mathrm{obs}$, comprised of already seen 100 workspaces with 200 unseen start and goal configurations in each workspace. The second test dataset, unseen-$X_\mathrm{obs}$, comprised of completely unseen 10 workspaces where each contained 2000 unseen start and goal configurations. In the Baxter experiments, we created a dataset comprised of ten challenging simulated environments, and we show the execution on the real robot. For each environment, we collected 900 paths for training and 100 paths for testing. The obstacle point clouds were obtained using a Kinect depth camera with the PCL \cite{Rusu_ICRA2011_PCL} and pcl\_ros\footnote{http://wiki.ros.org/pcl\_ros} package.      
\subsection{Models Architecture}
\subsubsection{Encoder Network}
For all environments except Baxter, we use encoder-decoder training, whereas for Baxter, we train the encoder and planning network end-to-end. Since the decoder is usually the inverse of the encoder, we only describe the encoder's structure. The encoding function $\mathrm{Enet}(\boldsymbol{x_\mathrm{obs}})$ comprised of three linear layers and an output layer, where each linear layer is followed by the Parametric Rectified Linear Unit (PReLU) \cite{trottier2016parametric}. The input to the encoder is a vector of point clouds of size $N_\mathrm{pc}\times d_w$ where $N_\mathrm{pc}$ is the number of data points, and $d_w \in \mathbb{N}$ is the dimension of a workspace.  
\subsubsection{Planning Network}
PNet is 9-layers and 12-layers DNN for Baxter and other environments, respectively. We use Parametric Rectified Linear Unit (PReLU) \cite{trottier2016parametric} for non-linearity. To add stochasticity, we Dropout (p) \cite{srivastava2014dropout} in all hidden layers except the last one. In point-mass and rigid-body cases, we fix the pretrained encoder parameters, since they were trained using the encoder-decoder method, and use them to compute environment encoding, whereas, for Baxter environments, we train the Enet and Pnet end-to-end. 
\begin{figure*}[h]
    \centering
    \begin{subfigure}[b]{0.23\textwidth}
       \includegraphics[height=3.9cm,width=4.2cm]{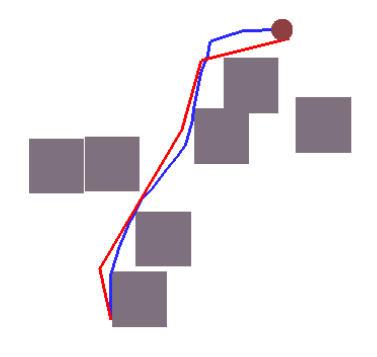}
        \caption{$t_\mathrm{R}=6.9s, t_\mathrm{MP}=0.50s$ }
    \end{subfigure}
    \begin{subfigure}[b]{0.23\textwidth}
       \includegraphics[height=3.9cm,width=4.2cm]{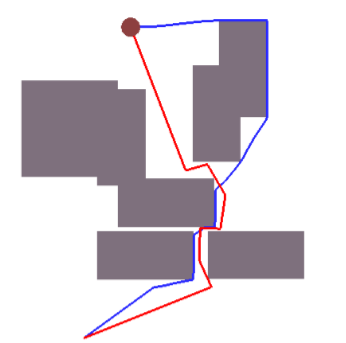}
        \caption{$t_\mathrm{R}=6.9s, t_\mathrm{MP}=0.50s$ }
    \end{subfigure}
    \begin{subfigure}[b]{0.23\textwidth}
       \includegraphics[height=3.9cm,width=4.2cm]{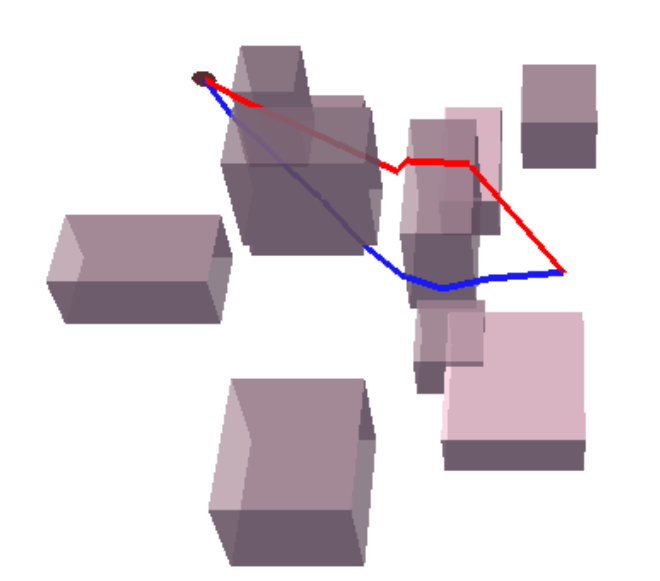}
        \caption{$t_\mathrm{R}=6.9s, t_\mathrm{MP}=0.50s$ }
    \end{subfigure}
    \begin{subfigure}[b]{0.23\textwidth}
       \includegraphics[height=3.9cm,width=4.2cm]{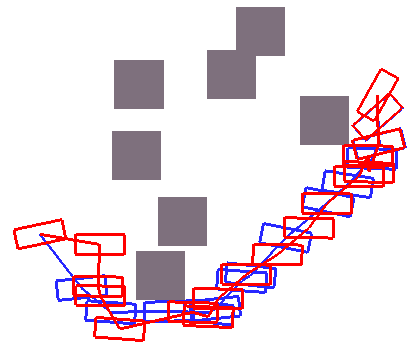}
        \caption{$t_\mathrm{R}=5.3s, t_\mathrm{MP}=0.44s$ }
    \end{subfigure}
    \caption{MPNet (Red) and RRT* (Blue) planning paths in complex 3D environments (c3D).}\label{c3D}
\end{figure*}

\begin{table*}
\vspace*{0.06in}
\centering 
\begin{tabular}{|c|c|c|c|c|c|c|c|}\hline
\multirow{2}{*}{Environment}& \multirow{2}{*}{Test case}&\multirow{2}{*}{ MPNet (NR)}& \multirow{2}{*}{MPNet (HR)} &  \multirow{2}{*}{Informed-RRT*}&\multirow{2}{*}{ BIT*} & \multirow{2}{*}{\scriptsize{$\cfrac{\mathrm{BIT}: t_\mathrm{mean}}{\mathrm{MPNet(NR)}:t_\mathrm{mean}}$}}   \\ 

&&& & & &   \\ \hline \hline 

\multirow{2}{*}{Simple 2D}& Seen $X_\mathrm{obs}$ & $0.11\pm0.037$& $0.19\pm0.14$ & $5.36\pm0.34$&$2.71\pm1.72$  & 24.64\\ \cline{2-7}
& Unseen $X_\mathrm{obs}$ & $0.11\pm0.038$& $0.34\pm0.21$ & $5.39\pm0.18$&$2.63\pm0.75$ & 23.91   \\ \hline

\multirow{2}{*}{Complex 2D}& Seen $X_\mathrm{obs}$ &$0.17\pm0.058$& $0.61\pm0.35$ & $6.18\pm1.63$& $3.77\pm1.62$ &22.17 \\ \cline{2-7}
& Unseen $X_\mathrm{obs}$ & $0.18\pm0.27$ & $0.68\pm0.41$ & $6.31\pm0.85$& $4.12\pm1.99$  & 22.89  \\ \hline

\multirow{2}{*}{Complex 3D}& Seen $X_\mathrm{obs}$ & $0.48\pm0.10$&$0.34\pm0.14$ & $14.92\pm5.39$& $8.57\pm4.65$ & 17.85 \\ \cline{2-7}
& Unseen $X_\mathrm{obs}$ & $0.44\pm0.107$& $0.55\pm0.22$ & $15.54\pm2.25$& $8.86\pm3.83$& 20.14  \\ \hline

\multirow{2}{*}{Rigid}& Seen $X_\mathrm{obs}$ & $0.32\pm0.28$& $1.92\pm1.30$ & $30.25\pm27.59$& $11.10\pm5.59$ & 34.69 \\ \cline{2-7}
& Unseen $X_\mathrm{obs}$ & $0.33\pm0.13$& $1.98\pm1.85$ & $30.38\pm12.34$& $11.91\pm5.34$& 36.09  \\ \hline

\end{tabular}
\caption{Time comparison of MPNet (NR: Neural Replanning; HR: Hybrid Replanning), Informed-RRT* and BIT* on two test datasets. Note in the right most column that MPNet is at least $20\times$ faster than BIT*.}
\vspace*{-0.2in}\end{table*}
%
%
%
%
%
%
%
%

\section{Results}
In this section, we compare the performance of MPNet with Neural-Replanning (MPNet: NR) and Hybrid-Replanning (MPNet: HR) against state-of-the-art motion planning methods, i.e., Informed-RRT* and BIT*, for the motion planning of the 2D/3D point-mass robots, rigid-body, and Baxter 7 DOF manipulator in the 2D and 3D environments. 

Figs. \ref{c3D} show different example scenarios where MPNet and expert planner, in this case RRT*, provided successful paths. The red and blue colored trajectories indicate the paths generated by MPNet and RRT*, respectively. The goal region is indicated as a brown colored disk. The mean computational time for the MPNet and RRT* is denoted as $t_\mathrm{MP}$ and $t_\mathrm{R}$, respectively. We see that MPNet is able to compute near-optimal paths for both point-mass and rigid-body robot in considerably less time than RRT*.   

Table I presents the CPU-time comparison of MPNet: NP and MPNet: HR against Informed-RRT* \cite{gammell2014informed} and BIT* \cite{gammell2015batch} over the two test datasets, i.e., seen-$X_\mathrm{obs}$ and unseen-$X_\mathrm{obs}$ . We report the mean times with standard deviation of all algorithms for finding the initial paths in a given problem. For initial paths, it is observed that on average the path lengths of benchmark methods were higher than the path lengths of MPNet. It can be seen that in all test cases, the mean computation time of MPNet with neural and hybrid replanning remained around 1 second. The mean computation time of Informed-RRT* and BIT* increases significantly as the dimensionality of planning problem is increased. Note that, on average, MPNet is about 40 and 20 times faster than Informed-RRT* and BIT*, respectively, in all test cases and consistently demonstrates low computational time irrespective of the dimensionality of the planning problem. In these experiments, the mean accuracy of MPNet: HR and MPNet: NP was 100\% and 97\% with the standard deviation of about 0.4\% over five different trials.

\begin{figure}[h]
\vspace*{0.05in}
    \centering
    \begin{subfigure}[b]{0.55\textwidth}
       \includegraphics[width=8.5cm]{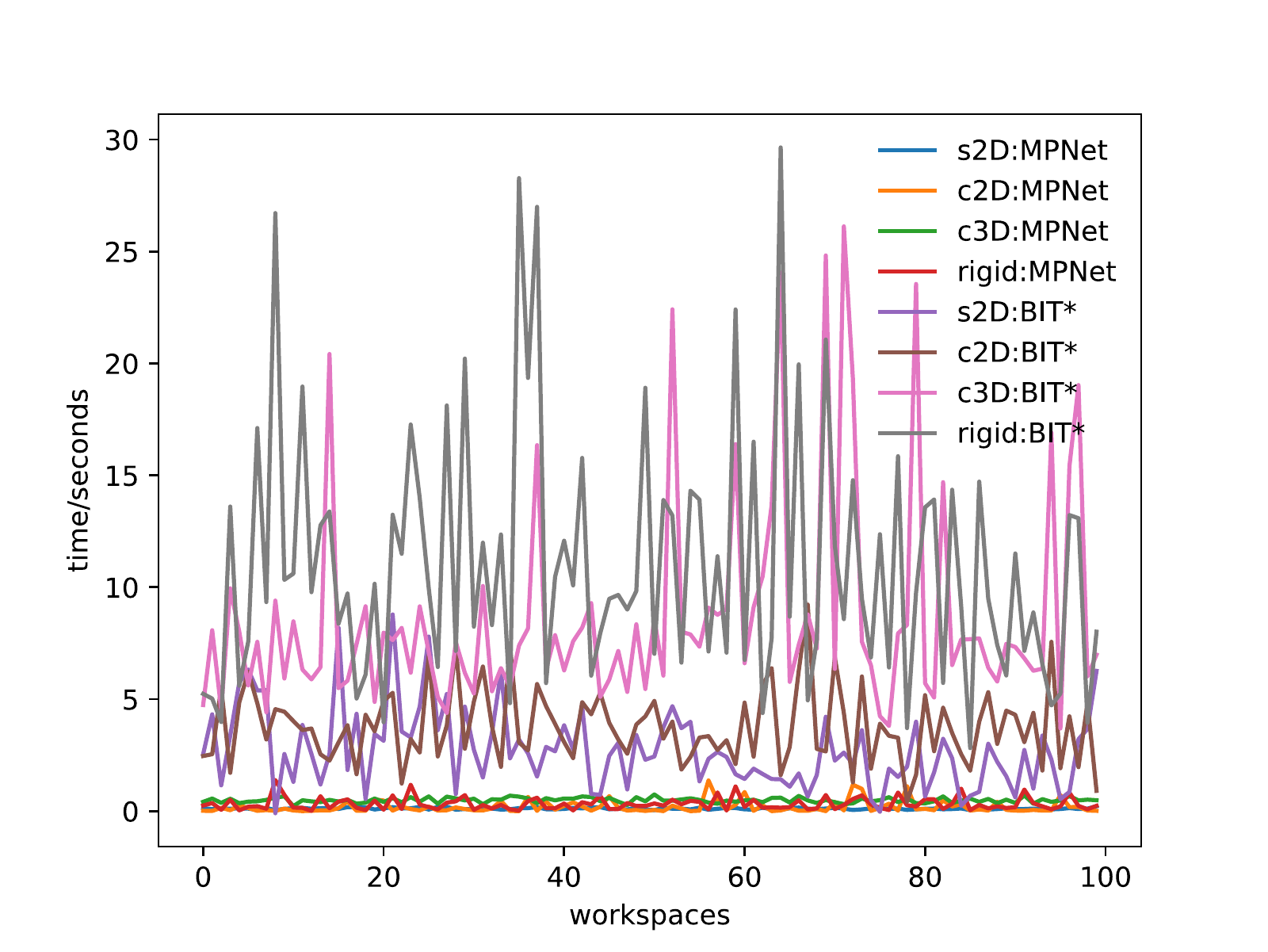}
        \caption{Test-case 1: seen-$X_\mathrm{obs}$}
    \end{subfigure}
        \begin{subfigure}[b]{0.55\textwidth}
       \includegraphics[width=8.5cm]{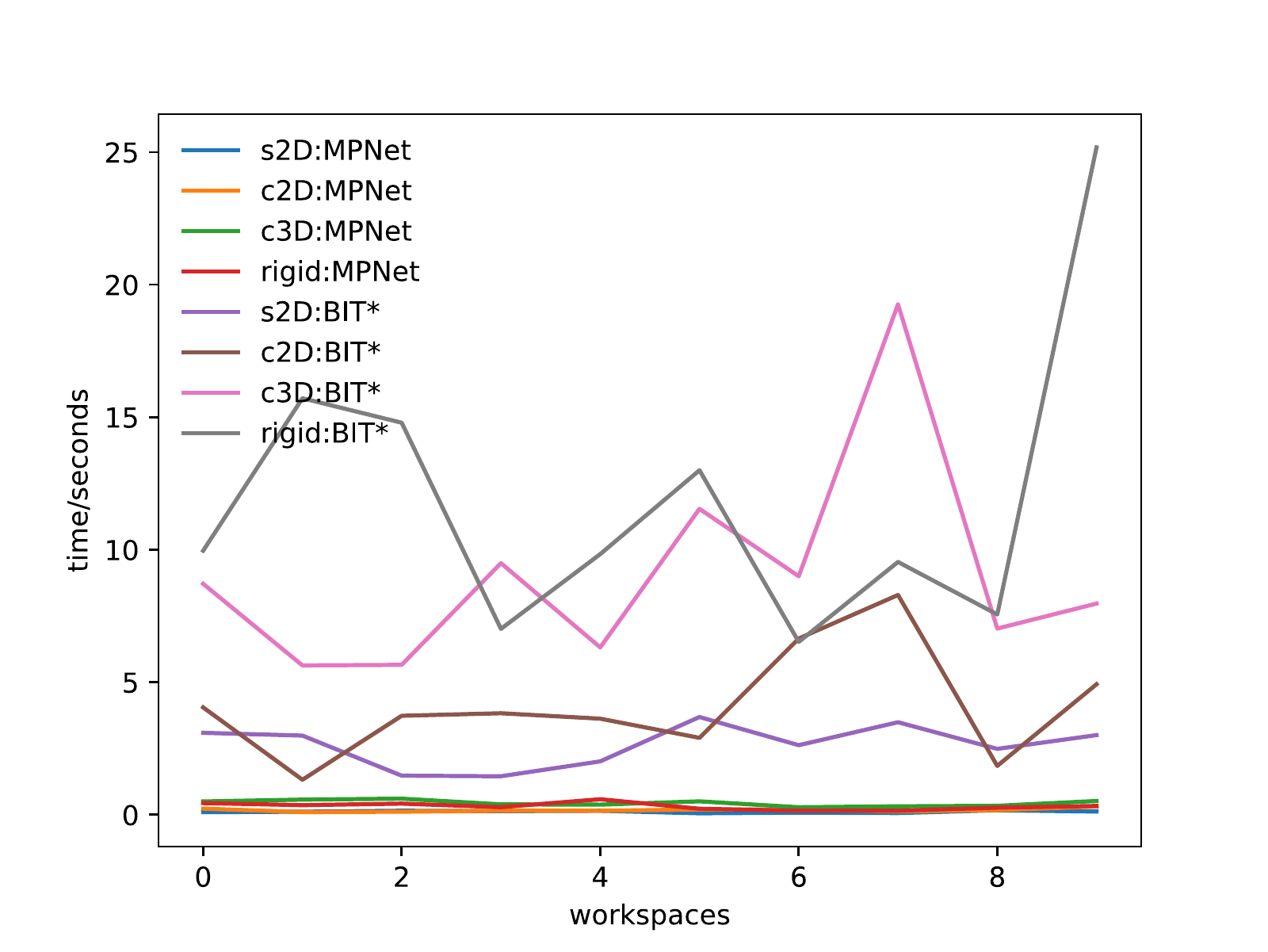}
    \caption{Test-case 2: unseen-$X_\mathrm{obs}$}
    \end{subfigure}
    \caption{Computational time comparison of MPNet and RRT* on test datasets. The plots show MPNet is more consistent and faster than BIT* in all test cases. }\label{tc}
\vspace*{-0.1in}\end{figure}

From experiments presented so far, it is evident that BIT*
outperforms Informed-RRT*, therefore, in the following experiments
only MPNet and BIT* are compared. Fig. \ref{tc} compares the mean computation time of MPNet: NP and BIT* in our two test datasets. It can be seen that the mean computation time of MPNet stays around 1 second irrespective of the planning problem dimensionality. Furthermore, the mean computational time of BIT* not only fluctuates but also increases significantly in the rigid-body planning problem.
Finally, Fig. \ref{baxter} shows  a single 7 DOF arm of the Baxter robot executing a motion planned by MPNet for a given start and goal configuration. In Fig. \ref{baxter}, the robotic manipulator is at the start configuration, and the shadowed region indicates the manipulator at the target configuration. On the Baxter's test dataset, MPNet took about 1 second on average with 85\% success rate. BIT* took about 9 seconds on average with a success rate of 56\% to find paths within a 40\% range of the path lengths found by MPNet, and was found to take up to several minutes to find paths within the 10\% range of average MPNet path lengths. 

\section{Discussion}
\subsection{Stochasticity through Dropout}
Our method uses Dropout \cite{srivastava2014dropout} during both online and offline execution. Dropout is applied layer-wise to the neural network and it drops each unit in the hidden layer with a probability $p \in [0,1]$, in our case $p=0.5$. The resulting neural network is a thinned network and is essentially different from the actual neural model \cite{srivastava2014dropout}.

\begin{figure}
    \centering
      \includegraphics[height=6.0cm]{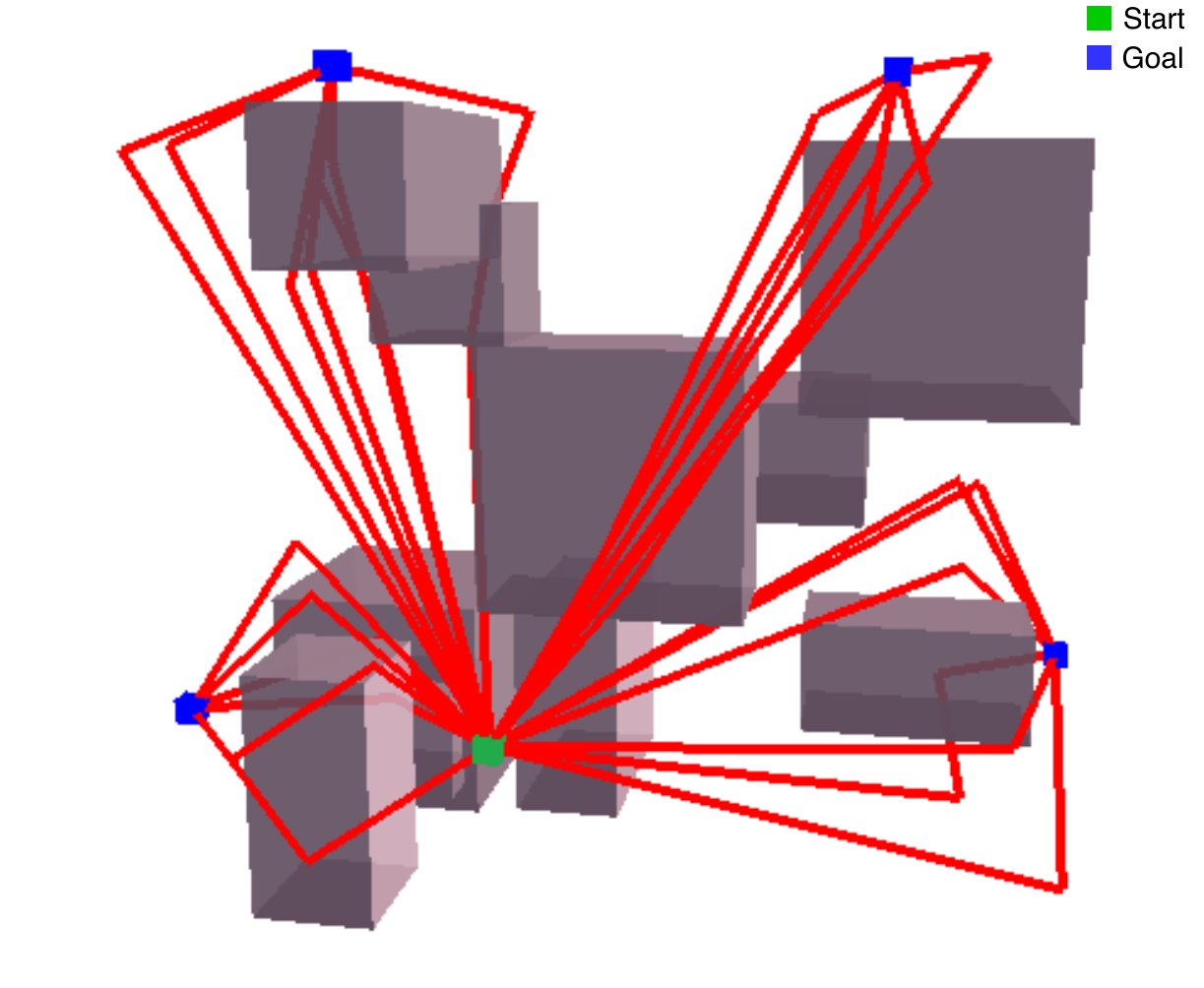}
    \caption{ MPNet generates multiple collision-free paths (red) between fixed start (green) and goal pairs (blue) in a finite-time due to its stochastic behavior.}\label{mpath}
\vspace*{-0.15in}\end{figure}

Note that, in the neural replanning phase, MPNet iterates over the non-connectable consecutive states of the coarse path to do motion planning on a finer level and thus, produces a new path. The replanning procedure is called recursively on each of its own newly generated paths until a feasible solution is found or a loop limit is reached. Dropout adds stochasticity to the Pnet which implies that on each replanning step, the Pnet would generate different paths from previous re-planning steps. This phenomenon is evident from Fig. \ref{mpath} where the Pnet generated different paths for a fixed start and goal configurations. These perturbations in generated paths for fixed start and goal help in recovery from the failure. Thus, adding Dropout increases the overall performance of MPNet. Furthermore, due to stochasticity by Dropout, our method can also be used to generate adaptive samples for sampling based motion planners \cite{qureshi2018deeply}. 
\subsection{Completeness}
In the proposed method, a coarse path is computed by a neural network. If a coarse path is found to be not fully connectable, a re-planning heuristic is executed to repair the non-connectable path segments to provide an end-to-end collision-free path. The completeness guarantees for the proposed method depends on the underline replanning heuristic. The classical motion planner based replanning methods are presented to guarantee the completeness of the proposed method. Since we use RRT*, our proposed method inherits the probabilistic completeness of RRTs and RRT* \cite{karaman2011sampling} while retaining the computational gains. 
\subsection{Computational Complexity}
This section formally highlights the computational complexity of the proposed method. Neural networks are known to have online execution complexity of $O(1)$. Therefore, the execution of lines 1-2 of Algorithm 1 will have a complexity no greater than $O(1)$. The lazy state contraction (LSC) heuristic is a simple path smoothing technique which can be executed in a fixed number of iteration as a feasible trajectory must have a finite length. Also, note that the LSC is not an essential component of the proposed method. Its inclusion helps to generate near-optimal paths. The computational complexity of the replanning heuristic depends on the motion planner used for replanning. We proposed the neural replanning and hybrid replanning methods. Since the neural replanner is executed for fixed number of steps, the complexity is $O(1)$. For the classical motion planner, we use RRT* which has $O(nlogn)$ complexity, where $n$ is the number of samples in the tree \cite{karaman2011sampling}. Hence, for hybrid replanning, we can conclude that the proposed method has a worst case complexity of $O(nlogn)$ and a best case complexity of $O(1)$. Note that, MPNet: NR is able to compute collision-free paths for more than 97\% of the cases, presented in Table I. Therefore, it can be said that MPNet will be operating with $O(1)$ most of the time except for nearly 3\% cases where the RRT* needs to be executed for a small segment of overall path given by MPNet:NR. This execution of RRT* on small segments of a global path reduces the complicated problem to a simple planning problem which makes the RRT* execution computationally acceptable and practically much less than $O(nlogn)$.
\section{Conclusions and Future work }
In this paper, we present a fast and efficient Neural Motion Planner called MPNet. MPNet consists of an encoder network that encodes the point cloud of a robot's surroundings into a latent space,, and a planning network that takes the environment encoding, and start and goal robotic configurations to output a collision-free feasible path connecting the given configurations. The proposed method (1) plans motions irrespective of the obstacles geometry, (2) demonstrates mean execution time of about 1 second in all presented experiments, (3) generalizes to new unseen obstacle locations, and (4) has completeness guarantees.

In our future work, we plan to extend MPNet to build learning-based actor-critic motion planning methods by combining it with proxy collision checkers such as the Fastron algorithm \cite{das2019fastron,das2017fastron}. Another interesting extension would be to address the challenge of kinodynamic motion planning in dynamically changing environments.




\bibliographystyle{IEEEtran}
\bibliography{references}
\nocite{*}
\end{document}